\documentclass[sigconf,anonymous=false]{acmart}
\pdfoutput=1

\usepackage{booktabs} 
\usepackage{color}
\usepackage{colortbl}
\usepackage{graphicx}

\definecolor{Gray}{gray}{0.93}
\newcolumntype{a}{>{\columncolor{Gray}}l}
\newcolumntype{b}{>{\columncolor{Gray}}r}

\usepackage{amssymb, amsmath, amstext}
\graphicspath{ {images/} }

\begin{document}
% Use this section to set the ACM copyright statement (e.g. for
% preprints).  Consult the conference website for the camera-ready
% copyright statement.

% Copyright
\copyrightyear{2020}
\acmYear{2020}
\setcopyright{acmcopyright}\acmConference[ETRA '20 Full Papers]{Symposium on Eye Tracking Research and Applications}{June 2--5, 2020}{Stuttgart, Germany}
\acmBooktitle{Symposium on Eye Tracking Research and Applications (ETRA '20 Full Papers), June 2--5, 2020, Stuttgart, Germany}
\acmPrice{15.00}
\acmDOI{10.1145/3379155.3391320}
\acmISBN{978-1-4503-7133-9/20/06}

% Use this command to override the default ACM copyright statement
% (e.g. for preprints).  Consult the conference website for the
% camera-ready copyright statement.

%% HOW TO OVERRIDE THE DEFAULT COPYRIGHT STRIP --
%% Please note you need to make sure the copy for your specific
%% license is used here!
% \toappear{
% Permission to make digital or hard copies of all or part of this work
% for personal or classroom use is granted without fee provided that
% copies are not made or distributed for profit or commercial advantage
% and that copies bear this notice and the full citation on the first
% page. Copyrights for components of this work owned by others than ACM
% must be honored. Abstracting with credit is permitted. To copy
% otherwise, or republish, to post on servers or to redistribute to
% lists, requires prior specific permission and/or a fee. Request
% permissions from \href{mailto:Permissions@acm.org}{Permissions@acm.org}. \\
% \emph{CHI '16},  May 07--12, 2016, San Jose, CA, USA \\
% ACM xxx-x-xxxx-xxxx-x/xx/xx\ldots \$15.00 \\
% DOI: \url{http://dx.doi.org/xx.xxxx/xxxxxxx.xxxxxxx}
% }

% Arabic page numbers for submission.  Remove this line to eliminate
% page numbers for the camera ready copy
% \pagenumbering{arabic}

\title[Deep semantic gaze embedding and scanpath comparison for expertise classification]{Deep semantic gaze embedding and scanpath comparison for expertise classification during OPT viewing}

\author{Nora Castner}
%\authornote{Presenter}
\affiliation{%
	\institution{Perception Engineering, University of T\"{u}bingen}
	\streetaddress{Sand 14}
	\city{T\"{u}bingen} 
	\state{Germany} 
	\postcode{72076}
}
\email{nora.castner@uni-tuebingen.de}

\author{Thomas K\"{u}bler}  
\authornote{Work of the authors is supported by the Institutional Strategy of the University of T\"{u}bingen (Deutsche Forschungsgemeinschaft, ZUK 63)}
\affiliation{%
	\institution{Perception Engineering, University of T\"{u}bingen}
	\streetaddress{Sand 14}
	\city{T\"{u}bingen} 
	\state{Germany} 
	\postcode{72076}}
\email{thomas.kuebler@uni-tuebingen.de}

\author{Katharina Scheiter} 
%\authornote{ADD}
\affiliation{%
	\institution{Leibniz-Institut f\"{u}r Wissensmedien}
	\streetaddress{Schleichstraße 6}
	\city{T\"{u}bingen} 
	\state{Germany}
}
\email{	k.scheiter@iwm-tuebingen.de}

\author{Juliane Richter}  
%\authornote{ADD}
\affiliation{%
	\institution{Leibniz-Institut f\"{u}r Wissensmedien}
	\streetaddress{Schleichstraße 6}
	\city{T\"{u}bingen} 
	\state{Germany}}
\email{	j.richter@iwm-tuebingen.de}

\author{Th\'{e}r\'{e}se Eder}  
%\authornote{ADD}
\affiliation{%
	\institution{Leibniz-Institut f\"{u}r Wissensmedien}
	\streetaddress{Schleichstraße 6}
	\city{T\"{u}bingen} 
	\state{Germany}}
\email{tf.eder@iwm-tuebingen.de}

\author{Fabian H\"{u}ttig}  
\authornote{Department of Prosthodontics}
\affiliation{%
	\institution{University Hospital T\"{u}bingen}
	\city{T\"{u}bingen} 
	\state{Germany} }
\email{	fabian.huettig@med.uni-tuebingen.de}

\author{Constanze Keutel}  
\authornote{Department of Radiology, Center of Dentistry, Oral Medicine and Maxillofacial Surgery}
\affiliation{%
	\institution{University Hospital T\"{u}bingen}
	\city{T\"{u}bingen} 
	\state{Germany}}
\email{constanze.keutel@med.uni-tuebingen.de}

\author{Enkelejda Kasneci}
%\authornote{ADDD.}
\affiliation{%
	\institution{Perception Engineering, University of T\"{u}bingen}
	\streetaddress{Sand 14}
	\city{T\"{u}bingen} 
	\state{Germany} 
	\postcode{72076}
}
\email{enkelejda.kasneci@uni-tuebingen.de}

\renewcommand{\shortauthors}{N. Castner et al.}

\begin{abstract}
 Modeling eye movement indicative of expertise behavior is decisive in user evaluation. However, it is indisputable that task semantics affect gaze behavior. %Thus, for interactive systems, user evaluation is two-fold: User expertise and scene information. 
 We present a novel approach to gaze scanpath comparison that incorporates convolutional neural networks (CNN) to process scene information at the fixation level. Image patches linked to respective fixations are used as input for a CNN and the resulting feature vectors provide the temporal and spatial gaze information necessary for scanpath similarity comparison. We evaluated our proposed approach on gaze data from expert and novice dentists interpreting dental radiographs using a local
 alignment similarity score. Our approach was capable of distinguishing experts from novices with 93\% accuracy while incorporating the image semantics. Moreover, our scanpath comparison using image patch features has the potential to incorporate task semantics from a variety of tasks. %reword, not confined to dental or medical expertise
\end{abstract}


% Author Keywords

\maketitle

\section{Introduction}
%%%%%%%%%%%%%%%%%%%%%%% Introduction

\begin{figure} %TODO:Get on first page, with all the authors
	\centering
	\includegraphics[width=0.45\textwidth]{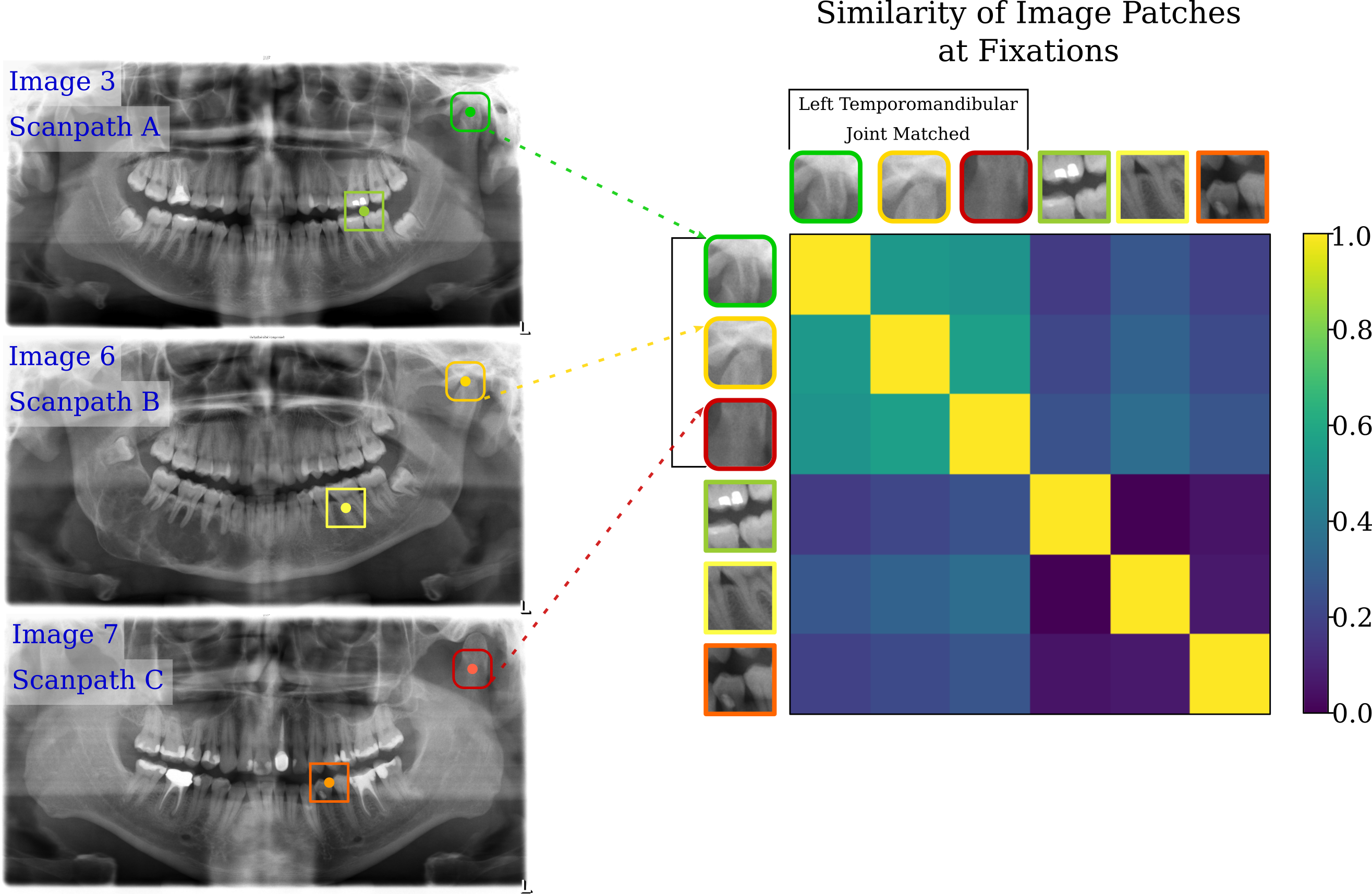}
	\caption{Matching image patch descriptors are recognized as similar across stimuli. When three different participants fixate on the left temporomandibular joint, the feature descriptors from DeepScan value them as similar. In contrast to when these participants fixate elsewhere, e.g. teeth, roots, etc.}~\label{fig:figureTeaser}
\end{figure}

Through eye movements, our thoughts, motivations, and expertise can be distinguished. We can accurately classify what someone is looking at and, more important, in what context they are looking at it, simply from the patterns in our gaze behavior. %TODO: rewrite   
Eye-tracking data is, however, still subject to large intra- and inter-individual variance. Neither two subjects are likely to look at a given stimulus in an identical way, nor is the same person likely to exhibit the identical gaze sequence when looking at the same stimulus twice. This variability becomes non-trivial when developing online systems that can recognize specific groups: e.g., distinguish experts from novices or doing performance prediction.
 
We measure these distinct gaze patterns as a scanpath: Areas of focus (fixations) where the eye behavior remains relatively still before moving to another area via a rapid eye movement (saccade)~\cite{holmqvist2011eye}. 
Discriminating scanpaths necessitates effective ways of aggregating and averaging gaze data over multiple trials to achieve converging summarizations of representative scanpaths (e.g. attention density maps~\cite{le2013methods}). 

Such aggregation techniques are simple to implement as long as subjects view the same stimulus from the same perspective, e.g., an image on a computer screen. Although, when  
either aggregation over a range of different stimuli or dynamic stimuli is required, analysis becomes challenging. 
For instance, semantically identical regions -- also known as areas of interest (AOIs) -- with regard to the studied task have to be identified and annotated. Once annotated, the sequence of AOIs visited by gaze can be analyzed as a proxy representation of the scanpath. 

However, even though it is apparent that task and subject differences affect scanpaths, often accurate prediction is still elusive. Greene et al.~\cite{greene2012reconsidering} failed to predict an observers' task from their gaze behavior using sequence information from manually defined AOIs. Additionally, when aggregating the scanpath data, ~\cite{borji2014defending} found they still could not accurately classify the task. Prediction increased in~\cite{kanan2014predicting} when considering the scanpath as a collection of features representing a fixations position and duration. Finally, the largest improvement in prediction performance was found when training an HMM model per stimulus~\cite{haji2014inverse}. Although it was accurate and incorporated spatial information, it has constrained applicability across stimuli.    

In order to apply task or subject prediction from scanpath information, conventional approaches that handle one image, one subject, or both are not feasible. One realm in particular that has shown promising potential for gaze behavior is training of medical personnel~\cite{van2017visual,gegenfurtner2011expertise,waite2019analysis}. For instance, gaze analysis has often been proposed as a measure for adaptive training systems (i.e. searching radiographs for pathologies~\cite{jarodzka2010convey,jarodzka2012conveying}, practicing surgery or laproscopy in VR/AR~\cite{law2004eye}). 
However, actually working training procedures are still scarce. Massed practice approaches, i.e., lengthy viewing of hundreds of radiographs, is still common educational practice~\cite{rozenshtein2016effect}. Even though it has been available for decades, as of now eye-tracking has yet to deliver the promises for adaptive training. The challenge of expediting a novice to expert solely through training gaze behavior has yet to be fully operational~\cite{van2017visual}.

In this work, we show how to incorporate high-level, deep neural network-generated image patch representations into classical scanpath comparison measures. We apply our method DeepScan to expertise classification on an eye movement dataset of expert and student dentists. Dentistry, in particular, relies heavily on effective visual inspection and interpretation of radiographs~\cite{huettig2014reporting}. Even then, panoramic dental radiographs are highly susceptible to diagnostic error~\cite{bruno2015understanding,akarslan2008comparison,douglass1986clinical,gelfand1983reliability}.
We demonstrate our method by decoding expertise from eye movements during dental radiograph inspection, which is a crucial first step towards adaptive learning procedures. It is worth noting, this metric is not confined to dental expertise recognition, rather developed with the intention for various use cases. It offers the future potential to assess student's learning progress in real-time and to adapt stimulus material based on current aptitude, while not being restricted to the stimulus material used during creation of the classifier.

\section{Revisiting Visual Scanpath Comparison}
%%%%%%%%%Related
\subsection{Traditional Approach: String Alignment}
One of the most common and traditional approaches to scanpath comparison is extraction of a similarity score by representing a scanpath as a sequence of symbols and comparing the resulting string to one another~\cite{anderson2015comparison}. AOIs on a given stimuli can be semantically or structurally linked to a symbol~\cite{cristino2010scanmatch,goldberg2010scanpath,jarodzka2010vector,kubler2014subsmatch}. Thus, coded strings provide information on the temporal and spatial order of the user's gaze behavior. Temporal resolution (i.e. fixation duration) can also be factored into the sequence~\cite{cristino2010scanmatch}. 

The output of such a comparison -- the similarity score -- is based on a total derived from rewarding matches and penalizing mismatches or gaps\footnote{inserting a space into one of the sequences.}. A scoring matrix can be used to represent the relative similarity of characters to one another~\cite{baichoo2017computational,day2010examining,goldberg2010scanpath}. A positive matching score represents similar regions and a negative score mismatches. Gaps are inserted in order to make neighboring characters match and to compensate small shifts of highly similar segments between the sequences. 

%TODO: can get rid of this papragraph and make it one sentence in the previous.
Global sequence alignment with a notion of AOI similarity can be performed via the Needleman-Wunsch algorithm~\cite{anderson2015comparison,needleman1970general}. Global sequence alignment determines the most optimal alignment for the entirety of two sequences. It has been shown to be a robust metric in scanpath comparison, e.g. in ScanMatch~\cite{cristino2010scanmatch}, classification of attentional disorder~\cite{galgani2009automatic}, multiple scanpath sequence alignment~\cite{burch2018eyemsa}, and expert and novice programmer classification~\cite{busjahn2015eye}. Castner et al.~\cite{castner2018scanpath} found incoming dental students with no prior training in radiograph interpretation could be classified from later semester students with 80\% accuracy from Needleman-Wunsch similarity scores.

Similarly, scanpath similarity from local sequence alignment has often been used as a robust classifier. Rather than deal with the entirety of two sequences, local alignment determines the most optimal aligned subsequence between the two. Local alignment compensates to a greater degree for sequences of differing lengths and is not as strongly influenced by differences in the beginning or end of the sequences ~\cite{khedher2018local}. 
For example, ~\cite{khedher2018local} used the Smith-Waterman algorithm~\cite{smith1981identification} for local alignment of medical undergrads' scanpaths during a clinical reasoning task. They found similarly well performing students had highly correlative scores. Similarly, ~\cite{ccoltekin2010exploring} found high comprehension and scanpath similarity of local subsequences in reading interactive map displays. 

Determining the optimal alignment between two sequences is computationally costly. 
Additionally, though commonly used, these methods suffer from a severe drawback: The manual selection of AOIs. This process is subjective, not only in which AOIs are considered relevant for the analysis, but also with regard to their size~\cite{cristino2010scanmatch,jarodzka2010vector}. For instance, Deitelhof et al.~\cite{deitelhoff2019influence} found that scanpath transitional information can be highly impacted by the AOI size and padding, which can affect validity. Moreover, some measures (e.g. Levenshtein distance) only rate exact matches and mismatches and do not consider any potential AOI similarity - and the ability of an algorithm to include this introduces the additional hard problem of judging AOI similarity objectively.%; as this would further complicate analysis.

Much of the prior literature on scanpath comparison using sequence alignment have employed manual AOI definitions. However, these approaches suffer errors in spatial resolution or require task-subjective AOI labels~\cite{jarodzka2010vector,cristino2010scanmatch}. K\"{u}bler et al.~\cite{kubler2014subsmatch} developed a method --SubsMatch-- for sequence comparison without AOI definitions, which uses a bag-of-words model and looks at the transitional behavior of subsequences. Castner et al.~\cite{castner2018scanpath} used these subsquence transitions from SubsMatch with an SVM Classifier~\cite{kubler2017subsmatch} and found comparable results to sequence alignment with grid AOIs.

However, these automatic approaches lack any notion of what is actually being looked at. Therefore, they usually perform excellent when subjects view the exact same stimulus (because then location identity corresponds to semantic identity to some extent). But when performing cross-stimulus analysis or the stimulus is subject to noise, performance drops significantly.

As of now, gaze pattern comparison is based either only on gaze location -- not on the semantic object that is being looked at -- or relies on human annotation to determine the semantics. 
Yet, scene semantics are absolutely critical for judging gaze behavior. For larger experiments and \emph{in the wild} head-mounted eye tracker data~\cite{wan2018fixation,pfeiffer2016eyesee3d}, manual annotation is unfeasible. We propose a method that combines the traditional approach of sequence alignment with deep learning for fixation target understanding. Combining these methods enables us to understand (and automatically analyze) the semantics behind a scanpath.

\subsection{Current Directions: Deep Learning}

Convolutional neural networks (CNNs) can provide information of image semantics that can be used for segmentation~\cite{long2015fully,chen2017deeplab} or classification~\cite{krizhevsky2012imagenet} and saliency prediction~\cite{huang2015salicon,hong2015online}, and many other applications. In the field of eye tracking research, they have also provided robust performance in eye movement behavior and scanpath generation~\cite{assens2017saltinet,liu2015predicting}.  For instance, methods using probabilistic models and deep learning techniques coupled with ground truth gaze behavior have been shown to predict fixation behavior ~\cite{kummerer2015information, wang2015deep}.

Concerning human scanpath classification, ~\cite{fuhl2019encodji} encoded gaze data as a compact image with the spatial, temporal, and connectivity represented as pixel values in the red, green, and blue channels respectively. These images were input for a CNN classifier, which showed high accuracy in classifying task-based gaze behavior. Mishra et al.~\cite{mishra2018automatic} followed a similar approach of depicting scanpath information as an image for a CNN sarcasm detector. 

Tao and Shyu~\cite{8794996} offer an approach similar to our proposed approach. They developed a CNN-Long Short Term Memory (LSTM) network that runs on scanpath-based patches from a saliency-predicted map\footnote{ASD specific saliency prediction from the Saliency4ASD challenge.} and classifies typical/autism spectrum disorder gaze behavior~\cite{8794996}. Square patches are defined based on fixation positions as they occur in the scanpath. Then, each patch is run through a shallow CNN, and the patch feature vector with the duration information provides an LSTM network input for classification from a dense layer from each patch~\cite{8794996}.      
Most notable, they maintain the sequential information of the scanpath.

We utilize powerful Deep Neural Network(DNN)-based feature descriptors to represent the semantics of a gaze sequence (scanpath). Our proposed approach follows a similar idea of incorporating the sequential fixation information in conjunction with visual features using a CNN. However, we extract scanpath similarity from the culmination of image patch features using the traditional approach of sequence alignment. For the current work, we chose local alignment in order to focus on common subsequences that could be indicative of expertise. Then, we cluster the scanpaths based on this similarity. Subsequently, we evaluate our proposed approach on detecting expert and student dentists' scanpaths when inspecting dental radiographs.

\section{Proposed Approach}
%%%%%%%%%%%%%%%%%%%%%%%Methodology
\begin{figure*}
	\centering
	\includegraphics[width=0.9\textwidth]{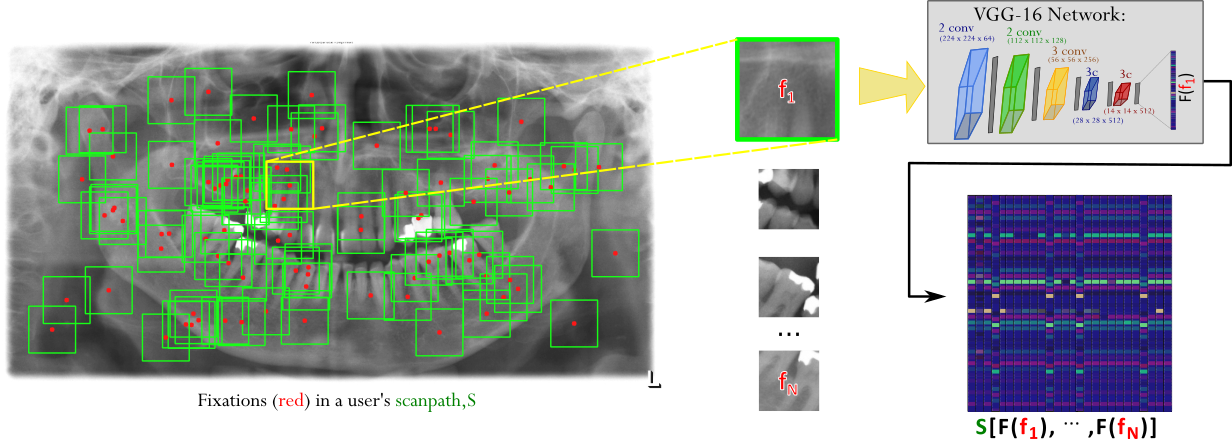}
	\caption{Proposed Model: DeepScan. For a scanpath, we extract the fixation locations and, using the VGG-16 CNN architecture, we create a feature corresponding to an image patch relative to the $i$th fixation F($f_{i}$). The resulting vector illustrating the scanpath $S$ can then be compared to another scanpath vector. In our work, we compared scanpaths via local alignment similarity. The pre-trained VGG-16 network consist of 5 blocks of convolutions with ReLus with max-pooling between each layer.}~\label{fig:figure1}
\end{figure*}

\subsection{Image Features at the Fixation Level}    
Each individual fixation corresponds to a visual intake of a certain stimulus region. We then encode each fixation location on the specific stimulus image by a vector that describes the local image region. We generate such encodings via the output from the VGG-16 architecture~\cite{simonyan2014very}. Accordingly, for each fixation location on the stimulus image, we extract a patch of $100\times100$ pixels as input to the network. This step is relatively similar to ~\cite{8794996}, although we determined that using a fixed size bounding box is adequate for our stimuli. The fixation coordinates indicate the center of the bounding box of the image patch, unless a fixation is too close to the stimulus borders. Then, appropriate shifting of the box along the x or y axis is necessary. 
 
The architecture we employed for patch processing originally takes $224 \times 224$ RGB input images. For the current evaluation on experts and students searching dental radiographs, our stimuli were grayscale with pixel dimensions $1920 \times 1080$. In development, we determined that patch sizes of $224 \times 224$ for our stimuli were too large (e.g. four or more teeth would be in this sized area). Smaller patches were more preferable so that enough information from an entity %region 
is extracted.
Therefore, we rescaled the $100 \times 100$ image patches to the desired input size for the network, and replicated the one channel image information to get three channels that can utilize the weights pre-trained on ImageNet~\cite{imagenet_cvpr09}. 

However, image patch input size and channels could be adapted for other stimuli or any other preferred network for the fixation encodings. The takeaway from this image patch approach is that through only the gaze: 1) we map the image features of interest in temporal order, and 2) we can extract the semantics from these features.

\subsection{CNN Architecture}
%16 stands for the number of weight layers in the network
For patch descriptor extraction, we employed a VGG-16 network~\cite{simonyan2014very} as implemented in keras\footnote{Python 3.6 with GPU compatibility.} and pre-trained on ImageNet. 
Figure \ref{fig:figure1} shows the network: Consisting of five blocks of convolutions, with each block followed by ReLUs and max-pooling. 

Since we are only interested in the features, we omit the fully-connected and prediction output layers of the model and use the output after max-pooling, which has $7 \times 7 \times 512$ dimensions, and flatten it to a $1 \times 1 \times 25 088$ vector. This feature description from the final convolutional layer, \textbf{F($f_{i}$)}, represents the image patch at the $i$th fixation $f_{i}$. 

The feature descriptors provide the semantic information for each fixation in a user's scanpath and are the equivalent to a symbol representation in the traditional string-sequence representation. In the following, we discuss the changes required in the alignment algorithm in order to work with alignment scores generated by comparing these image features to each other. Figure \ref{fig:figureTeaser} shows an example of how similar features can compare to each other.  

We chose the VGG-16, in contrast to a network pre-trained on radiology images since it is more generalizable to a variety of tasks and stimuli. Additionally, it is publicly available and easily integrated for replication purposes.  
Pre-trained networks for medical images are often not publicly available due to the data sensitivity and protection, and any existing architectures for these images are not yet up to par with the generic image trained architectures.    
Choosing a network that is trained for a specific stimulus category, e.g., panoramic radiographs or other X-Rays, might improve results. However, it introduces the risk of limiting data analysis to specific elements, which is comparable to manual AOI selection. Ultimately, though our approach is evaluated on medical image expertise, we developed it for generalizability in multiple applications.  %for cross-application/task use 

\subsection{Local Alignment}

Once we have descriptors for each fixation, we assemble them as a scanpath.
The resulting matrix of image features at each fixation creates a scanpath matrix. 
$S_{A} = (F_{f_{1}},F_{f_{2}},\cdots,F_{f_{N}})$. With this matrix representation, we can compare its similarity to the matrix representing another scanpath. 

For scanpath comparison, we perform local alignment using a variant of the Smith-Waterman Algorithm. We preferred local alignment scoring over global alignment due to its ability to find similar subsequences, even if the scanpaths may otherwise be highly varying~\cite{khedher2018local}. Moreover, we did not want to enforce strict global alignment due to different viewing times required by students and experts.
In sequence alignment, the penalty system can have a major effect on values in the scoring matrix, and therefore, the similarity score~\cite{baichoo2017computational}. Our scoring choice prioritizes finding long rather than short similar subsequences by accumulating scores. Equation \ref{eq:score} details the scoring system used for the current evaluation:
\begin{equation}
M_{ij} = \max 
\begin{cases}
M_{i-1,j-1} + \ c - \displaystyle \sum_{i,j} |A_{:F_{j}} - B_{:F_{i}}| , & \text{\small{Match}} \\
M_{i,j-1} - gap, & \text{\small{Gap  in  A}}\\
M_{i-1,j} - gap, & \text{\small{Gap  in  B}}\\
0 & \text{\small{No Similarity}}.
\end{cases}
\label{eq:score}
\end{equation}

Where $M$ is the scoring matrix of size $(n+1) \times (m+1)$ for two scanpaths\ $A$ and $B$ with $n$ and $m$ fixations respectively. Element $M_{i,j}$ takes the maximum value based on if there is a match between the values at index $j$ of scanpath $A$ and index $i$ of scanpath $B$. The original algorithm scores matches as the score value added to the value at the previous indices: $M_{i-1,j-1} + score(a_{j},b_{i})$. Then, if there is no match, it determines whether
the value of a gap penalty ($gap$) in either scanpath, or no similarity ($0$) are more optimal for the score.

The interesting part of our approach is contained in the calculation of the match score. Since it is highly unlikely that two features will be exactly the same, we cannot explicitly match or mismatch. Therefore, we calculate the score by taking the sum of absolute differences in feature descriptor $j$ of scanpath $A$ and descriptor $i$ of scanpath $B$ as shown in the first line of equation \ref{eq:score}. This is simple to implement and cheap to compute, but other metrics such as cosine or Euclidean distance could also be used. This procedure leads to a dissimilarity score between the image patches. The more dissimilar the image patches, the larger the scoring value. 

In order to convert it to a similarity score, we can subtract the dissimilarity score from a constant $c$.
We calculated $c$ in equation \ref{eq:score} by averaging the sum of the differences for all features between all scanpaths of one random image in the dataset.
Therefore, $c$ was $21,049$ in the evaluation of our proposed approach. 
This constant affects highly similar image patches positively, but highly dissimilar image patches are penalized negatively with the same weight. Meaning it functions similar to a match/mismatch threshold.
Additionally, we set our gap penalty (lines 2 and 3 in eq.\ref{eq:score}) to $42,000$ to highly penalize gaps, therefore almost double $c$. 

\begin{figure}
	\centering
	\includegraphics[width=0.33\textwidth]{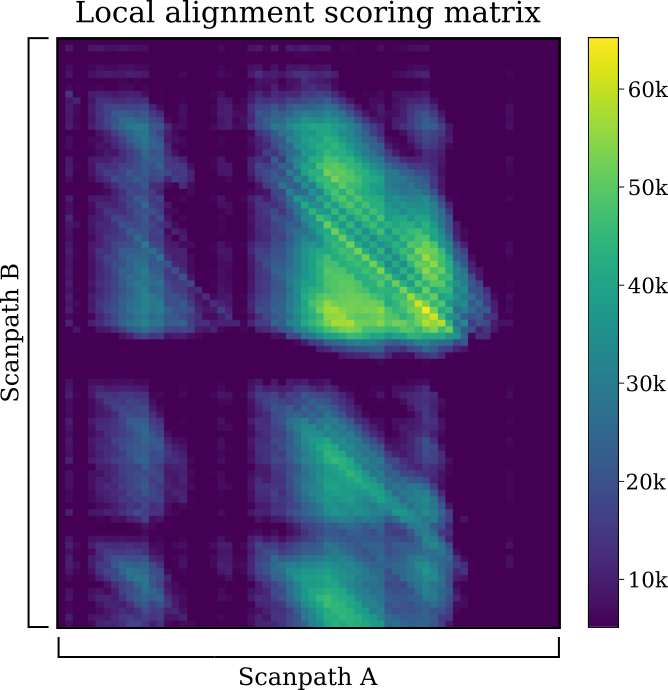}
	\caption{Scoring matrix of the local alignment. Backtracing from the index with the highest value (yellow) will give you the optimal local alignment of two scanpaths.}~\label{fig:figure2}
\end{figure}

This choice of $c$ makes the algorithm consider about half of the image patches relatively dissimilar to each other. Furthermore, gaps are penalized quite strongly, resulting in compact alignments that are not drastically influenced by large differences in sequence lengths. Figure \ref{fig:figure2} shows an example of the similarity matrix created from the local
alignment performed for two scanpaths. The maximum value in the matrix is the similarity score~\cite{smith1981identification}.
In figure \ref{fig:figure2}, the highest yellow color indicates the final similarity score and backtracing from this index till $0$ will give the optimal local alignment of both sequences.

The resulting similarity score for the two scanpaths is $\max(M)$. Then, we normalize this score based on the length of the shorter scanpath, thus:

\begin{equation}
similarity = \dfrac{\max(M)}{\min{(|S_{A}|,|S_{B}|)}}.
\label{eq:score2}
\end{equation}

%\textbf{"The expectation score is defined as the average score that the scoring system (substitution matrix and gap penalties) would yield for a random sequence"- from wiki}   

We compared the performance of our DeepScan method to Smith-Waterman local alignment of hand-labeled semantic AOIs (the gold standard in adding semantic information to image patches, see Supplementary Material Figure 1). These AOIs indicate specific anatomical structures and regions across the dental radiographs and provide the paramount in semantic information that can be represented in a scanpath. For scoring the semantic scanpath comparisons, we used a simple, standard scoring system: $1$ for matches, $-1$ for mismatches, and $-2$ for gaps.

\section{Evaluation}
%%%%%%%%%%%results
\subsection{Scanpath Data of Dentists}
\label{subsec:participants}
Students (n=$57$) were incoming dental students (sixth semester) from their initial pre-med studies. %17 + 24 + 16 = 57
They had no prior training in dental radiograph interpretation, but basic conceptual knowledge in general medical concepts. Experts (n=$30$, average $10.16$ years experience) were dentists working in the local university clinic with all the proper qualifications and some had further licensing for other particular specializations (e.g. Endontology, Prosthetics, Orthodontology, etc.). Diagnostic performance results from both groups indicated that the experts had $79.91\%$ higher pathology detection accuracy than students \footnote{Performance metrics and expert results can further be found in \cite{castner2018overlooking}}. %Students' recall: 27.92, Experts': 50.23. 

Both students and experts were asked to perform a visual search task of panoramic dental radiographs (OPTs); then following image inspection, indicate any areas indicative of pathologies. Students had 90 seconds to inspect each OPT, where experts had 45 seconds to inspect each OPT. This shortened duration was due to the research indicating that experts are much faster when visually inspecting radiographs~\cite{gegenfurtner2011expertise,turgeon2016influence}. Students inspected two blocks of 10 OPTs in one experimental run and experts -- due to their hard-pressed schedules -- inspected 15 OPTs. 

All eye tracking data was collected with SMI RED250 remote eye trackers sampling at 250Hz attached to laptops with FullHD displays. A quality assessed calibration\footnote{less than one degree average deviation from a four point validation.} was performed for each participant before and during data collection. 
Gaze data, i.e. fixations, were determined using a velocity based metric provided by the eye tracker's software. Further details of the data collection and pre-processing can be found in ~\cite{castner2018scanpath,castner2018overlooking}. 

For compatibility, we chose to evaluate gaze data from the first 45 seconds of each student participant, in line with the experts' total viewing time. Additionally, our model is only evaluated on gaze data for the 10 OPTs that both groups viewed. Gaze data was lost for two expert participants due to software failure. Also, $5$ participants were excluded due to having high data loss (under $80\%$ tracking ratio\footnote{A metric reported from SMI indicating proportion of valid gaze signals.} and $3$ or more low signal quality images) leaving $25$ experts and $54$ students for the final analysis. The resulting total for all participants for all images was 733 scanpaths. 

\subsection{Similarity Scoring}
\begin{figure}
	%\centering
	\includegraphics[width=0.5\textwidth]{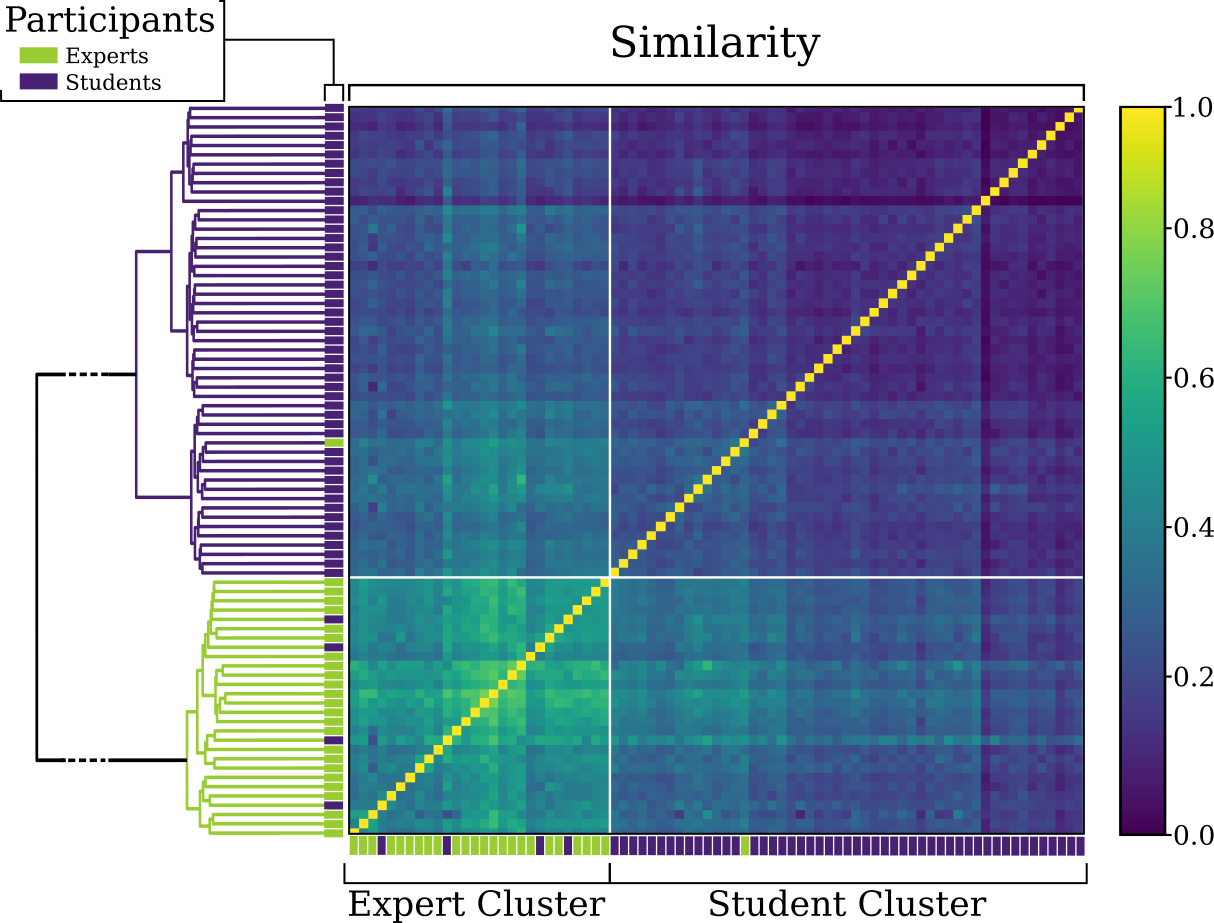}
	\caption{Similarity matrix of subjects' average scanpath behavior. Purple labels indicate students' gaze behavior. Green labels indicate experts' gaze behavior. Values closer to yellow indicate higher similarity, Where the diagonal is a participant compared against themselves. Values shown on the diagonal are rescaled relative to values off-diagonal solely for perceivability. On the y-axis is the resulting clustering of the dendrogram, which recognized 2 clusters. On cluster (purple) with mainly students and the other cluster (green) with mainly experts. }~\label{fig:figure3}
\end{figure}

We performed local alignment of the scanpath vectors with patch features for each participant for all images. In order to get the scanpath behavior representative of each participant, we averaged a participants' similarity output for all images. Figure \ref{fig:figure3} shows the similarity scores from DeepScan of each participants' scanpath behavior over the images viewed in pairwise comparison to other participants.
The diagonal of the matrix indicates the highest similarity value, which is a participants' gaze behavior compared to his or herself. 

From the similarity matrix, a trend is apparent where experts (labeled green in figure \ref{fig:figure3}) show higher similarity scores among each other, as visible by the more yellow values. Conversely, students' gaze behavior shows less similarity among each other, especially when compared to experts.

\subsection{Hierarchical Clustering}
We clustered the similarity scores of all participants using agglomerative hierarchical clustering~\cite{johnson1967hierarchical,corpet1988multiple,west2006eyepatterns}. As the similarity matrix can easily be inverted to a distance matrix, the unsupervised clustering approach was straight forward; however one could introduce additional weighting factors or more complex classification methods on top as well. 
This approach evaluates the distance between data points and links closer in distance clusters until one cluster remains~\cite{johnson1967hierarchical}. Partitioning the clusters then is determined by the linkage distance. We used Ward's~\cite{johnson1967hierarchical} method for proximity definition, which minimizes the sum of the squared distances of points from the cluster centroid. 

\paragraph{Average Gaze Behavior of Each Subject}

For the scores of each student and expert summed over all images, the resulting dendrogram  (2-dimensional tree view of the nested clusters) is shown on the y-axis in figure \ref{fig:figure3}.
%The distance between clusters is the sum of squared differences within all clusters

The clustering seen in figure \ref{fig:figure3} recognizes two main clusters evident in the gaze behavior with the majority of students in one cluster (purple cluster, purple labels) and the majority of experts (green cluster, green labels) in the other. Table \ref{tab:table1} calculates the true positive rate (TPR) when utilizing the clustering as a classification for both students and experts as well as the overall accuracy. We achieved $93.7\%$ accuracy. We also found two clusters evident in the traditional local alignment with manual AOIs; however more students were misplaced in the expert cluster (as seen in table \ref{tab:table1}), resulting in an overall accuracy of $85\%$.

\begin{table}
	
	\centering
	\begin{tabular}{l l a l a c c}
		% \toprule	
			& \multicolumn{2}{c}{{\small \textit{Student}}}
		& \multicolumn{2}{c}{{\small \textit{Expert}}}
		&  \multicolumn{2}{c}{{\small \textbf{\textit{Accuracy}}}}  \\
		&{\footnotesize \textit{Feature}}&\cellcolor{white}{\footnotesize \textit{Semantic}}
		&{\footnotesize \textit{Feature}}&\cellcolor{white}{\footnotesize \textit{Semantic}}
		&{\footnotesize \textit{Feature}}&{\footnotesize \textit{Semantic}}\\
		\midrule
		{\small \textit{Student}} & 50 & 44 & 1 & 1 & & \\
		{\small \textit{Expert}} & 4 & 10 & 24 & 24 & & \\
		\midrule
		{\small \textit{TPR}} & 92.5\,\% & 81.5\,\% & 96.0\,\% & 96.0\,\% &\textbf{93.7\,\%} & \cellcolor{Gray}86.06\,\% \\
		\bottomrule
	\end{tabular}
	\caption{Performance of linkage clustering for our approach (\emph{Feature}) and Semantic AOIs as measured by the True Positive Rate (TPR). Two main clusters were found based upon the gaze behavior for both approaches.}~\label{tab:table1}
\end{table}

\paragraph{Gaze Behavior on the Image Level}
We then ran the hierarchical clustering for participants' gaze at the image level (over all 733 datasets and not the average similarities for each participant as above). 
The dendrogram also recognized two clusters, therefore we calculated the number of experts in one cluster and the number of students in the other. The achieved accuracy for our approach was $68.62\%$: Experts had $85.65\%$ TPR and students had $61.18\%$ TPR.  The achieved accuracy for the traditional, semantic approach was $64.39\%$: Experts had $51.76\%$ TPR and students had $93.27\%$ TPR.
This slight dip in performance could be attributed to pathology differences in the stimuli. Previous literature has also found that gaze behavior of expert and novice dentists can be highly stimulus dependent, where dental radiographs considered easy to interpret evoke similar gaze behavior in experts and novices~\cite{turgeon2016influence,grunheid2013visual}.

\subsection{Cross-Image Classification}
To further see whether we could predict classification performance on an image level, we performed a leave one subject and one image out cross-validation using the similarity scores from DeepScan. We performed classification to 1) see whether we could predict a participant's expertise from their scanpath on a new image, not contained in the set that we compare to. 2) to confirm that certain stimuli may affect the similarities more than others.
For each subject, we ran a 3-nearest neighbor classifier, trained on the remaining subjects and images. 
Table \ref{tab:table2} shows the performance for each image. Here, it is clear that for some images, distinguishing expert and student scanpaths becomes more difficult. For instance, image 1 shows a heavy tendency to classify all participants' scanpaths for that image as experts, and image 3 shows a tendency to over-classify as students. %TODO: have fabian explain why?
% TKnote: This could be used to indicate which stimulus material actually invokes a learning effect and therefore should be included in the training and which could probably be left out. When using a real classifier instead of the clustering approach, we could "learn" these nonsense stimuli and therefore push classification accuracy even futher.
Nevertheless, five images allowed us to determine expertise of a new subject on a new stimulus that were not contained in the data we used for the classification. %NC: Image 5 was ALMOST at acceptable performance!!!!!!
% TKnote: this means, we could use it in an online setting with new students and even change/update the stimulus material without changing the classification
Especially, image 8 shows the highest accuracy in classifying level of expertise, meaning this OPT and its semantics can possibly trigger experts to inspect the image in a distinctive way.   
   
The cross-validation for the traditional local alignment scoring for the scanpaths with manual AOIs, showed better performance on the image level than DeepScan, and slightly better overall ($77\%$ versus $73\%$ respectively). Thereby, it is possible that we cannot yet utilize the full potential of semantic encoding using the feature approach. However, given that DeepScan is purely data driven, its results are comparable and relegates the tedious process of manual AOI labeling. Retraining the network on OPT data might help the encoding to come closer to manually-defined semantic labels.

%TODO: 1 decimal place
\begin{table}
	\scalebox{0.822}{
		\centering
		\begin{tabular}{l r a r a r l b a}
			% \toprule	
			& \multicolumn{2}{c}{{\small \textit{Expert TPR}}}
			& \multicolumn{2}{c}{{\small \textit{Student TPR}}} &
			\multicolumn{4}{c}{\textbf{\textit{Accuracy}}}\\
			
			& {\footnotesize \textit{Feature}} & \cellcolor{white}{\footnotesize \textit{Semantic}} &
			{\footnotesize \textit{Feature}} & \cellcolor{white}{\footnotesize \textit{Semantic}} &  \multicolumn{2}{c}{{\footnotesize \textit{Feature}}} & \multicolumn{2}{c}{{\footnotesize \textit{Semantic}}} \\
			
			\midrule
			{\small \textbf{Chance:}} & \multicolumn{2}{c}{32\,\%} & \multicolumn{2}{c}{68\,\%} & 
			{\small \textit{Overall}} & {\small $\kappa$} &
			\cellcolor{white}{\small \textit{Overall}} & \cellcolor{white}{\small $\kappa$} 
			\\
			\midrule
			{\small \textit{Image 1}} & 100\,\% & 75\,\% & 20.4\,\% & 76.6\,\% & 44.9\,\% & 0.14 & \textbf{78.2\,\%}  & 0.52\\
			{\small \textit{Image 2}} & 59.1\,\% & 68.2\,\% & 83.3\,\% & 85.4\,\% &\textbf{75.7\,\%}& 0.43 & \textbf{80\,\%} & 0.54\\
			{\small \textit{Image 3}} & 28.6\,\% & 66.7\,\% & 93.5\,\% & 80.4\,\% &73.1\,\% &0.26 &\textbf{76.1\,\%} &  0.46\\
			{\small \textit{Image 4}} & 52.4\,\% & 57.1\,\% &89.8\,\% & 83.7\,\% &\textbf{78.6\,\%} &0.45 & \textbf{75.7\,\%}& 0.41\\
			{\small \textit{Image 5}} & 76.2\,\% & 53.4\,\% &68.6\,\% & 88.2\,\% &70.8\,\%&0.39 & \textbf{77.8\,\%} &0.43\\
			{\small \textit{Image 6}} & 66.7\,\% &75\,\% &67.9\,\% & 81.1\,\% &65.5\,\%&0.31 & \textbf{79.2\,\%}  & 0.54\\
			{\small \textit{Image 7}} & 60.9\,\% & 30.4\,\% &86.5\,\% &90.4\,\% &\textbf{78.7\,\%}&0.49 & 72\,\% &0.24\\
			{\small \textit{Image 8}} & 73.9\,\% &91.3\,\% &88.2\,\% &68.6\,\% &\textbf{83.8\,\%}&0.62 &\textbf{75.7\,\%} & 0.51 \\
			{\small \textit{Image 9}} & 45.8\,\% &58.3\,\% &92.6\,\% &96.3\,\% &\textbf{78.2\,\%}&0.43 & \textbf{84.6\,\%}&0.60\\
			{\small \textit{Image 10}} & 30\,\% &80\,\% &96.2\,\% &65.4\,\% &77.8\,\%&0.32  & 69.4\,\% &0.37\\
			\midrule
			{\small \textit{Overall}} & 60.1\,\%&65.5\,\% & 78.2\,\%&82\,\% & 72.7\,\%& 0.37& \textbf{76.9\,\%} &0.46 \\
			\bottomrule
	\end{tabular}}
	\caption{Performance of kNN classifier when one image is left out and each participants' expertise for that image is predicted. Note that chance level is not $50\%$, therefore we provide Cohen's Kappa ($\kappa$) as an indicator of performance, with bold text indicating fair performance.}\label{tab:table2}
	% A Cohen's Kappa value above 0.4 is considered fair performance, whereas above 0.75 would be considered very good.
\end{table}

\begin{figure} %TODO: if time, recolor to imply duration
	\centering
	\includegraphics[width=0.4\textwidth]{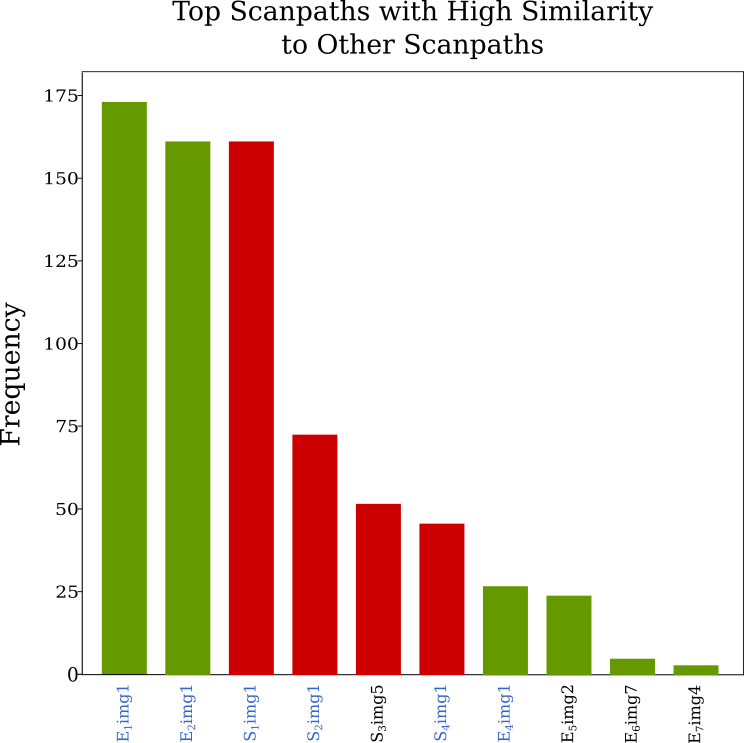}
	\caption{The top scanpaths who have the highest frequencies of similarities to other scanpaths; With experts indicated in green and students indicated in red. The majority of theses scanpaths are for image 1, as indicated by the blue text.}~\label{fig:figure6}
\end{figure}

Additionally, we sorted the similarity scores of all scanpaths from DeepScan to isolate those that expose especially high similarity values to many other scanpaths. We hoped to extract archetype-scanpaths this way. The histogram in figure \ref{fig:figure6} shows that two expert scanpaths had the highest similarity scores to the most other scanpaths. Interestingly enough, both these scanpaths and a number of the other high similarity scanpaths are for image 1. %they are similar to 45.8\% of the data sets
% TKnote: and that's one of those that make distinction between experts and novices hard - therefore it is quite logical that it scores good towards many others.
Thus from the local alignment similarity, certain scanpaths from image 1 offer highly similar subsequences to other scanpaths regardless of image. Image 1 was one of the stimuli that made a distinction between expertise levels hard. It might therefore represent a standard scanpath for checking OPTs that abstracts over special attributes of individual stimuli.

The two experts scanpaths (illustrated by their image patches) with the most highest similarities to each other and many other subjects' scanpaths are shown in figure 2 in the Supplementary Material.

\section{Discussion}
%%%%%%%%%%%%%%%%%%%%%%%%Discussion
We were able to successfully extract similarities in the scanpath behaviors between experts and the differences towards students gaze behavior while interpreting panoramic dental radiographs. Our developed scanpath comparison approach uses temporal scanpath information to extract image features at the fixation level. The resulting similarity comparison of scanpaths therefore incorporates this image information into the traditional approach of sequence alignment to detect patterns between the behaviors.
	
\begin{figure}
	\centering
	\includegraphics[width=0.45\textwidth]{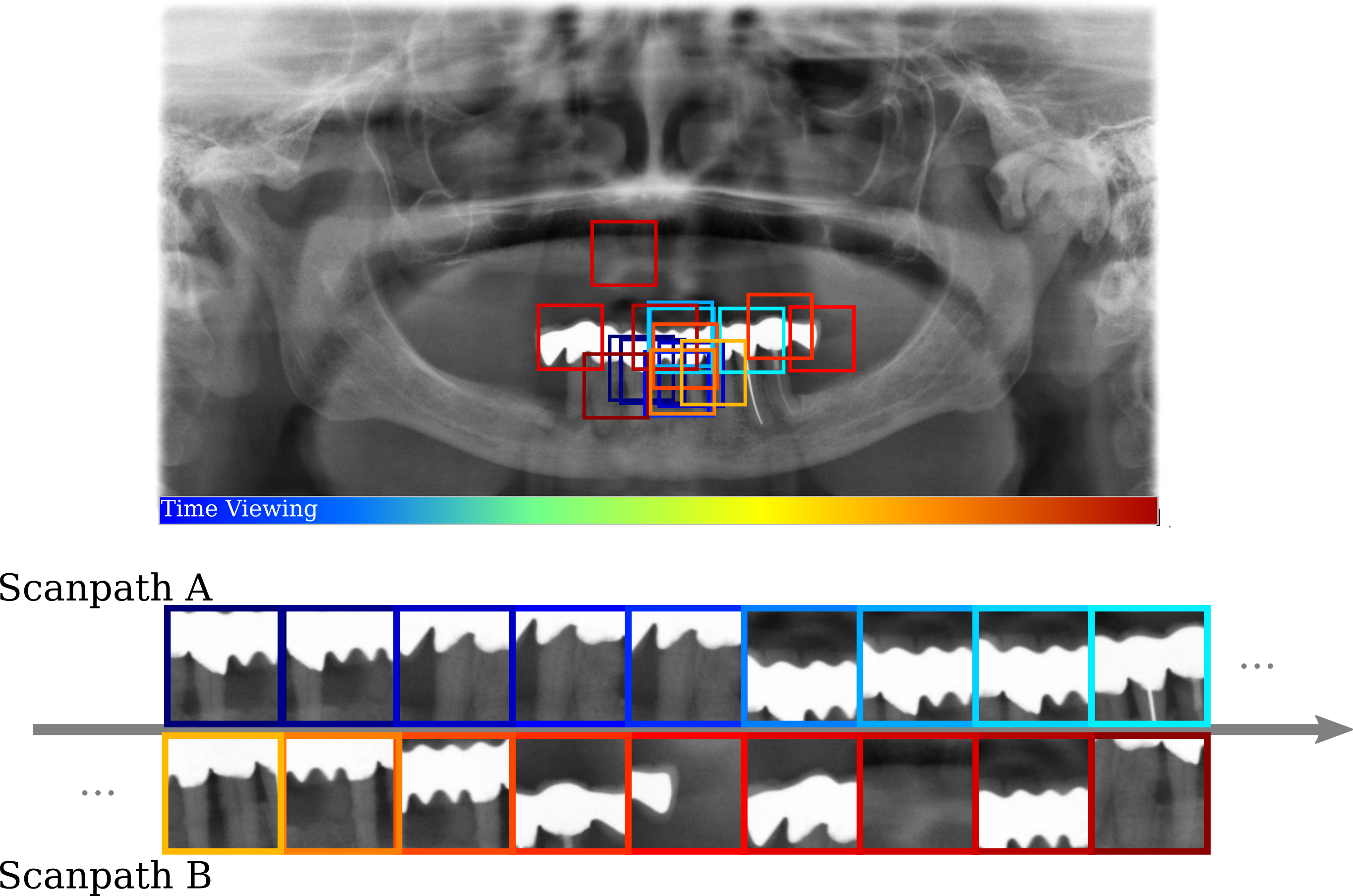}
	\caption{Two relatively dissimilar scanpaths from students. The local alignment finds the optimal matching subsequence starting in scanpath A at the twentieth fixation (far left top) and in scanpath B at the fiftieth fixation (far left bottom).}~\label{fig:figure7}
\end{figure}

From traditional local alignment techniques using image features, we found that experts showed highly similar behavior to each other and therefore, were more likely to be clustered together. More interesting, students' similarity scores indicated that their scanpaths were not highly similar to those of experts, but also there was no distinct homogeneity among themselves (see figure \ref{fig:figure7}). One possible reason for their low similarity to each other could be that they are incoming students with some conceptual background; however, they had no training on radiograph interpretation. Previous research has found that students evoke more systematic search strategies after training, resulting in more similar gaze behaviors~\cite{kok2016systematic,van2017visual}. Additionally, the heterogeneity of background and training can affect scanpath similarity~\cite{davies2016exploring}. Possibly students have varying levels of conceptual knowledge or familiarity with radiographs before entering their first year of dental studies.

Our algorithm was able to accurately classify unseen scanpaths given scanpaths from other participants and other images. Although we found that, depending on the image, it could be easier or harder to differentiate the levels of expertise from the scanpath similarities.  
This finding is, however, in alignment with previous studies specifically on dentists and dental radiograph examination. For instance, ~\cite{turgeon2016influence} found that radiographs defined as easy to interpret offered no differences in the gaze behavior of experts and novices. Castner et al. \cite{castner2018overlooking} also found that even among experts some images evoked highly differing gaze behavior to achieve accurate anomaly detections. 
%NC: a bit of a hard trasition to we found image differences, but our scanpath metric offers inter images classification....

With the system at hand, we could classify expertise of dentist students in an adaptive feedback setting from viewing just a single stimulus (with decent accuracy), even if the stimulus itself is an arbitrary OPT that is unknown to the classifier. This could be used to guide students through the learning process and to adapt the difficulty of stimulus material to their current knowledge level. When viewing multiple stimuli (which students do in the current mass practice approach), classification accuracy can be increased.

Futhermore, we observed that some stimuli allowed for a classification of expertise, while others did not. We could utilize this information as a hint on which stimuli are likely to induce a training effect and to differentiate them from stimuli that are too easy (for the current student).

Moreover, we designed DeepScan to handle image variability. One image feature descriptor of a patch in one image can match to similar patches in other images (see figure \ref{fig:figureTeaser}); This way, scanpaths can be more easily compared cross-stimuli, but this process also replaces a manual AOI-annotation. By the assumption that similar semantic meaning in a visual task corresponds to similar looking features in the stimulus, we have introduced a notion of stimulus semantics into the automated scanpath interpretation. A similar workflow could be used to compare data where the annotation of dynamic AOIs is usually unfeasible, e.g., recordings of mobile eye-tracking devices to each other. Furthermore, we do not restrict the algorithm to individual annotated AOIs, but represent each fixation by its feature descriptor, no matter whether a data analyst would deem it relevant for the analysis at hand or not.

One limitation for the current work could be the methodological confound of the viewing time differences in the expert and student paradigms. Since a consistently longer viewing time for the students would heavily affect the similarity scoring regardless of normalization, we took the first 45 seconds of the students, so that our similarity scores would be less biased by their longer scanpaths.

\section{Conclusion}
%%%%%%%%%%%%%%%%%%%%%%%%%%Conclusion
Our proposed model for scanpath classification, DeepScan, is capable of extracting gaze behavior indicative of expertise in dental radiograph inspection. More important, this approach employs deep learning to extract image features. Consequently, human expert gaze behavior coupled with relevant image semantic extraction offers an accurate approach to automated scanpath classification. However, the motivation for this model does not finish here. Rather, it was developed for applicability not only in the medical expertise domain, but also for scenarios with dynamic, semantically varying tasks (i.e. Training in VR, real world scenarios with mobile eye tracking).

Future directions of the proposed approach optimization for online classification of scanpaths. We chose a local alignment evaluation as a traditional approach to scanpath comparison, since it provides for a standard and robust evaluation of the scanpath feature matrix created. DeepScan has the potential for online use and further evaluation are therefore necessary for working towards integrating this model into adaptive feedback scenarios.

%\section{Acknowledgments}
%We thank all the volunteers, and all publications support
%and staff, who wrote and provided helpful comments on previous
%versions of this document.
%Author 2 gratefuly acknowledges the support from [BLANK] %
%the Institutional Strategy of the University of Tübingen (Deutsche Forschungsgemeinschaft, ZUK 63)

% REFERENCES FORMAT
% References must be the same font size as other body text.
\bibliographystyle{ACM-Reference-Format}
\bibliography{sample}

\end{document}